\documentclass[journal]{IEEEtran}
\usepackage{amsmath,amsfonts}
\usepackage{algorithmic}
\usepackage{algorithm}
\usepackage{array}
\usepackage{textcomp}
\usepackage{stfloats}
\usepackage{url}
\usepackage{verbatim}
\usepackage{graphicx}
\usepackage{caption}
\usepackage{subcaption}
\usepackage{cite}

\begin{document}

\title{Visual Place Recognition for Large-Scale UAV Applications}
\author{Ioannis Tsampikos Papapetros*, Ioannis Kansizoglou, and Antonios Gasteratos%
\thanks{ I.T. Papapetros (*corresponding author), I. Kansizoglou, and A. Gasteratos are with the Department of Production and Management Engineering, Democritus University of Thrace, Xanthi, Greece. 
        {\tt\small ipapapet@pme.duth.gr, ikansizo@pme.duth.gr, agaster@pme.duth.gr}}}

\markboth{Journal of \LaTeX\ Class Files,~Vol.~XX, No.~X, X~XXXX}%
{Shell \MakeLowercase{\textit{et al.}}: A Sample Article Using IEEEtran.cls for IEEE Journals}


\maketitle

\begin{abstract}
Visual Place Recognition (vPR) plays a crucial role in Unmanned Aerial Vehicle (UAV) navigation, enabling robust localization across diverse environments. Despite significant advancements, aerial vPR faces unique challenges due to the limited availability of large-scale, high-altitude datasets, which limits model generalization, along with the inherent rotational ambiguity in UAV imagery.
To address these challenges, we introduce LASED, a large-scale aerial dataset with approximately one million images, systematically sampled from 170,000 unique locations throughout Estonia over a decade, offering extensive geographic and temporal diversity. Its structured design ensures clear place separation significantly enhancing model training for aerial scenarios. Furthermore, we propose the integration of steerable Convolutional Neural Networks (CNNs) to explicitly handle rotational variance, leveraging their inherent rotational equivariance to produce robust, orientation-invariant feature representations.
Our extensive benchmarking demonstrates that models trained on LASED achieve significantly higher recall compared to those trained on smaller, less diverse datasets, highlighting the benefits of extensive geographic coverage and temporal diversity. Moreover, steerable CNNs effectively address rotational ambiguity inherent in aerial imagery, consistently outperforming conventional convolutional architectures, achieving on average 12\% recall improvement over the best-performing non-steerable network. By combining structured, large-scale datasets with rotation-equivariant neural networks, our approach significantly enhances model robustness and generalization for aerial vPR. 
\end{abstract}


\section{Introduction}

The ability to determine the precise position of mobile platforms is crucial across a wide range of applications, from urban navigation and virtual reality to enabling full autonomy in robotics. While numerous methods have been developed to tackle this problem, each comes with specific strengths and limitations. With the rapid increase in Unmanned Aerial Vehicle (UAV) applications~\cite{teixeira_survey_2023,papapetros_multi-layer_2020,mitroudas_light-weight_2024,radoglou-grammatikis_compilation_2020}, the need for accurate and efficient positioning solutions has become even more pronounced, particularly in expansive and dynamic environments. 
As the demand for autonomy in critical applications grows, integrating diverse information sources becomes essential to enhance system capabilities and ensure reliability. 

Traditional localization methods, though widely used, can sometimes fall short under various operational constraints. High-resolution cameras, by capturing rich visual data, offer a robust alternative that can complement these methods or can be used as a standalone solution when all other methods fail. For high-altitude UAVs, cameras are particularly advantageous as they provide broad-area coverage, deliver detailed environmental context, and support high-precision localization through visual cues. This capability is especially valuable in environments where traditional sensors might be limited by range, resolution, or availability of infrastructure.

In a typical visual localization pipeline, visual data is processed through three main steps: constructing a comprehensive reference map, establishing initial visual correspondences between the input data and the map, and refining these correspondences to enhance positional accuracy~\cite{toft_long-term_2022,sarlin_coarse_2019,taira_inloc_2021,torii_are_2021}. This process leverages the detailed visual cues captured by cameras to determine the platform’s location, rendering it a versatile approach for environments where traditional methods face limitations. 
Within this pipeline, Visual Place Recognition (vPR) is crucial for matching new visual inputs to the reference map by determining if two images depict the same location~\cite{tsintotas_revisiting_2022,lowry_visual_2016}. This step is essential for enabling robust and scalable localization, particularly across extensive and diverse environments, making vPR foundational for effective vision based navigation.

Until recently, the demand for vPR systems has been primarily driven by ground-based applications, such as augmented and virtual reality, autonomous driving, and mobile robotics.
While such systems have achieved remarkable accuracy and scalability, largely thanks to neural networks trained on extensive datasets, they face significant challenges when adapted to aerial applications. State-of-the-art ground-based vPR methods often depend on vast datasets of terrestrial imagery, optimized for recognizing features at street level~\cite{berton_rethinking_2022,ali-bey_gsv-cities_2022,warburg_mapillary_2020,arandjelovic_netvlad_2016}. Many approaches build upon these pre-trained networks by incorporating specific logic or use-case assumptions to enhance their performance~\cite{keetha_anyloc_2024,papapetros_visual_2023,papapetros_semantic-based_2023,chen_geocluster_2024}.
However, the aerial vPR domain lacks large-scale, high-altitude datasets, resulting in methods that rely on pre-trained networks not optimized for aerial imagery. These networks struggle with the unique characteristics of aerial data, including drastic viewpoint changes, significant scale variations, and diverse environmental conditions. Consequently, current aerial vPR methods often fall short in performance and efficiency.
The absence of domain-specific training data means that aerial vPR systems cannot fully leverage the advancements made in ground-based vPR. To overcome these challenges, it is crucial to develop and utilize datasets and networks specifically tailored to high-altitude imagery. 

To address the unique challenges of aerial vPR and bridge the gap between ground-based and UAV-based localization methods, this work makes the following key contributions:

\begin{itemize}
    \item LASED: A Large-Scale Aerial vPR Dataset – We introduce LASED, the first publicly available, high-altitude vPR dataset with ~1M images spanning over a decade, covering diverse geographic regions, seasonal changes, and long-term environmental transformations. This dataset provides a foundation for developing and evaluating aerial vPR methods at scale.
    \footnote{Available at: https://github.com/ipapap/Visual-Place-Recognition-for-Large-Scale-UAV-Applications}
    
    \item Integration of Steerable Convolutional Networks \cite{freeman_design_1991,weiler_learning_2018} for Aerial vPR – We propose the first application of steerable Convolutional Neural Networks (CNNs) in aerial vPR, addressing rotational ambiguity inherent in UAV imagery. Our experiments demonstrate significant improvements in performance compared to conventional architectures.
    
    \item Extensive Benchmarking of vPR Methods – We thoroughly evaluate state-of-the-art vPR approaches in the aerial domain, highlighting performance trends across multiple datasets and identifying key limitations of ground-trained models when applied to aerial scenarios.
    
    \item Generalization and Training Data Impact Analysis – We assess how the scale and diversity of training data affect model generalization, comparing performance across different datasets and demonstrating the importance of dataset structure in aerial vPR.
\end{itemize}

\section{Related Works}

Visual place recognition has been proven essential for many tasks, in both robotics, and computer vision, with a key objective to determine if two images represent the same place ~\cite{cadena_past_2016}. Many works employ methods that are based on well-defined rules~\cite{calonder_brief_2010,dalal_histograms_2005} that leverage information, such as colors, textures and shapes, to extract meaningful information from an image in order to uniquely describe the depicted place, showing promising results mainly on trivial scenes~\cite{zaffar_cohog_2020,papapetros_visual_2022}. However, contemporary challenges, such as large viewpoint variances as well as visual changes from lighting conditions or seasonal changes, have pushed the field to adopt neural networks to extract that kind of information~\cite{masone_survey_2021}. These networks, when trained on large amount of data, are able to exploit higher-level visual information to describe the content of an image, leading to significant improvements in recognition performance. 
Early works utilize CNNs~\cite{gu_recent_2018,moutsis_evaluating_2023} trained on large generic datasets, such as ImageNet~\cite{krizhevsky_imagenet_2012} to encode the visual data of the captured places, showing improved performance even under challenging environmental conditions~\cite{suenderhauf_place_2015}. 

Utilizing the ample street-level imagery, several large-scale datasets emerge, tailored specifically for the vPR task~\cite{warburg_mapillary_2020,arandjelovic_netvlad_2016,berton_rethinking_2022,ali-bey_gsv-cities_2022}. Scaling from thousands to millions of images from mostly urban or suburban places, can provide the neural networks with enough data to achieve excellent performance on a diverse set of environments, even under significant changes in visual and environmental conditions. While the availability of extensive datasets plays a major role in achieving high performance, advancements in training schemes, network architectures and the problem-specific strategies employed by various methods have also been essential in improving results.


Most notably, the NetVLAD~\cite{arandjelovic_netvlad_2016} training scheme, utilizes a weakly-supervised learning strategy to train the underlying model from multiple panoramic images of different places, lacking the orientation information. Thus, correspondences between different observations cannot be reliably established, making it difficult to ensure that same-place observations contain overlapping content, while different-place observations remain distinct. This results in noisy annotations and ultimately limits performance. More recent datasets~\cite{warburg_mapillary_2020,berton_rethinking_2022,ali-bey_gsv-cities_2022}, offer high ground-truth accuracy allowing for greater discrimination between places, offering a great boost in performance. Some methods~\cite{berton_rethinking_2022,berton_eigenplaces_2023} propose training strategies that split the training dataset in discrete smaller groups and use those for iteratively training the proposed network, guaranteeing that for each group, same-place observations share visual content while cross-place images do not. Different works~\cite{ali-bey_gsv-cities_2022} introduce data, sampled in well-defined regions without any overlap, avoiding complex data organization methods, while others utilize the spatial proximity of the observations to weight the training loss accordingly~\cite{leyva-vallina_generalized_2023}. In this work, similar to~\cite{ali-bey_gsv-cities_2022}, we ensure that different places pose no overlap with each other. Unlike ground-based imagery, where large spatial separation is necessary to ensure no overlap, especially of distant objects, in aerial views observations span is limited, thus no large gaps are necessary among neighboring places. In such a way, no complex data organization strategies are necessary, simplifying the overall training procedure while still maximizing the number of places per spatial region.  


\begin{table*}[t]
\centering
\caption{Comparison of available aerial datasets. Scale denotes the total number of images in each dataset, as their primary objectives vary and may not be exclusively tailored for vPR. Longitudinal Samples indicates the number of distinct temporal observations per location, where 1+1 refers to virtual views generated from the same source.}
\resizebox{\textwidth}{!}{%
\begin{tabular}{|l|c|c|c|c|c|}
\hline
\textbf{Name}            & \textbf{Scale (\#Frames)} & \textbf{Altitude}       & \textbf{Geographic Coverage} & \textbf{Longitudinal Samples} & \textbf{Diversity} \\ \hline
\textbf{VPAIR}~\cite{schleiss_vpair_2022}           & 15k                   & High                     & Medium                         & 2                  & Urban, Grasslands, Forests \\ \hline
\textbf{ALTO}~\cite{cisneros_alto_2022}            & 40k                      & High                   & Medium                       & 2                  & Urban, Rural, Grasslands, Forests,  Rivers, Lakes \\ \hline
\textbf{SUES-200}~\cite{zhu_sues-200_2023}        & 80k                    &  Medium, High                      & Low                          & 1+1       & Urban                   \\ \hline
\textbf{AerialVL}~\cite{he_aerialvl_2024}        & 30k                       & Medium, High                       & Medium                         & 2     & Urban, Rural, Rivers, Lakes      \\ \hline
\textbf{University-1652}~\cite{zheng_university-1652_2020}        & 150k        & Ground, Medium, High              & Medium                          & 1+1   & Urban              \\ \hline
\textbf{UAV-VisLoc}~\cite{xu_uav-visloc_2024}        & 7k                       & High, Very High                       & Low                          & 2         & Urban               \\ \hline
\textbf{DenseUAV}~\cite{dai_vision-based_2024}        & 27k                       & Low                       & Low                          & 3                        & Urban               \\ \hline
\textbf{LASED (Ours)}    & 1M                  & High                   & High                         & 10                     & Urban, Rural, Grasslands, Forests, Rivers, Lakes,  Coastal    \\ \hline
\textbf{LASED-Test (Ours)} & 60k              & High                   & Medium                       & 2           & Urban, Rural, Grasslands, Forests, Mountainous, Rivers, Lakes           \\ \hline
\end{tabular}%
}
\label{tab:datasets}
\end{table*}

Ground-based vPR methods have benefited immensely from the availability of large-scale datasets. However, despite being optimized specifically for the vPR task, these datasets differ drastically in viewpoint and visual content compared to the aerial perspectives encountered in UAV applications. Furthermore, aerial vPR introduces unique challenges due to the operational characteristics of aerial platforms~\cite{zaffar_are_2019,moskalenko_visual_2025}. Large altitude variations, in-plane rotations, and lack of structured environment are a few of those problems. While some datasets attempt to address these challenges, they often fall short in providing the scale or diversity necessary for comprehensive model development and evaluation. Table \ref{tab:datasets} provides an overview of the currently available popular aerial datasets. Although some datasets tailored for the aerial domain have been published in recent years~\cite{xu_uav-visloc_2024,dai_vision-based_2024,zheng_university-1652_2020,schleiss_vpair_2022,he_aerialvl_2024,cisneros_alto_2022,zhu_sues-200_2023}, most are relatively small, with total image number well bellow $100,000$, featuring a very limited number of different locations. In contrast, ground-based datasets often contain millions to tens of millions of images~\cite{berton_rethinking_2022}, offering far greater diversity and scale. While many aerial datasets capture diverse environments and trajectories, spanning up to a few hundred kilometers across neighboring cities, they often lack temporal variety, with most focusing on captures taken close in time. This limits their ability to support the development of models that are robust to seasonal, weather-related, or long-term environmental changes.

However, aerial applications, can span a wide range of operational altitudes, from just above the ground to as high as 100 kilometers above sea level. In this work, inspired by~\cite{schleiss_vpair_2022}, we define ``high altitude'' loosely, as the range of 200-400 meters above ground level and focus on this.

Large-scale, well-structured datasets have been instrumental in driving advancements in ground-based vPR, where many state-of-the-art solutions rely on pre-trained models built on extensive, diverse datasets~\cite{papapetros_semantic-based_2023,garg_delta_2020,papapetros_visual_2023}. By adopting similar principles, aerial vPR can benefit from the scale and structure needed to train robust models. This motivates the creation of datasets like LASED, which combine geographic diversity and temporal depth to provide a solid foundation for UAV-based tasks.

In addition to the data itself, training strategies play a crucial role in vPR performance. Most of the mentioned datasets rely on a query/reference structure, which, while effective for evaluation, can limit the flexibility of advanced training approaches. Ground-based state-of-the-art solutions have shown that carefully structuring datasets to ensure clear place separation and minimize overlap can significantly boost performance~\cite{ali-bey_gsv-cities_2022,berton_rethinking_2022}. Applying these strategies to aerial datasets can enhance their utility, paving the way for models that are both more robust and better equipped to handle the unique challenges of aerial applications.

LASED follows this approach by incorporating approximately 1 million images with a decade of temporal coverage while maintaining a structured design inspired by gsv-cities~\cite{ali-bey_gsv-cities_2022}. Its architecture ensures clear place separation and minimizes overlap, supporting modern training strategies~\cite{kansizoglou_deep_2022,wang_multi-similarity_2019,balntas_learning_2016,wang_cosface_2018}. These characteristics make it a valuable resource for advancing aerial vPR, enabling models to generalize more effectively across diverse environmental conditions.

While high-quality data are crucial for training effective models, it is ultimately the neural network architecture that encodes visual information into useful representations. Recent research suggests splitting this process into two primary modules: the backbone and the aggregator~\cite{ali-bey_gsv-cities_2022}. The backbone is responsible for extracting relevant features from the image, capturing essential visual information. The aggregator, on the other hand, combines these features into a compact, meaningful representation that can be used for recognition tasks. This modular design allows for flexibility; the backbone~\cite{krizhevsky_imagenet_2012,he_deep_2016,simonyan_very_2015} can be updated with state-of-the-art models as they emerge, without altering the carefully optimized aggregator~\cite{ali-bey_gsv-cities_2022,berton_rethinking_2022,radenovic_fine-tuning_2019}. Such a framework ensures adaptability and maintains the robustness of the overall system. 

Recent works in the aerial domain have introduced innovative approaches for improving geo-localization by segmenting key regions within images and aligning them across views~\cite{dai_transformer-based_2022}, while others have explored leveraging contextual scene information to enhance matching performance~\cite{wang_each_2022}. These methods primarily focus on cross-view geo-localization, where the goal is to recognize locations despite extreme viewpoint variations, such as matching UAV imagery with satellite or ground-level references. While these approaches are valuable for bridging large perspective gaps, they fall outside the scope of this work, which concentrates on high-altitude UAV-based vPR under consistent aerial viewpoints. 

In this work, we focus on ResNet-based~\cite{he_deep_2016} backbone architectures, as they are widely adopted in state-of-the-art methodologies, and evaluate their performance with different aggregation methods to identify the most effective solutions for aerial vPR tasks. Additionally, inspired by the unique challenges of the aerial domain—particularly the unknown rotation of UAVs relative to the ground—we propose, for the first time, the integration of steerable convolutional neural networks~\cite{weiler_learning_2018,freeman_design_1991} into the vPR pipeline to address rotational invariance, increasing the robustness of the produced representations and thus vPR performance.

\section{Dataset}
The effectiveness of neural networks in vPR relies heavily on the availability of high-quality training data. Existing aerial datasets, as discussed earlier, often lack the scale, diversity, and structure required to train models capable of addressing the unique challenges of aerial vPR. To fill this gap, we introduce LASED (LArge-Scale Estonia Dataset), a dataset specifically designed to support the training of robust and versatile aerial vPR models. 

LASED comprises approximately 1 million images captured over a decade, featuring over 170 thousand unique locations, encompassing diverse environments across the entire country of Estonia. Its multi-season imagery and decade-long temporal span provide the variety needed to train models resilient to environmental changes, such as seasonal variations and long-term transformations. The dataset is carefully structured to ensure clear place separation and avoid overlap, enabling the application of advanced training strategies commonly used in state-of-the-art ground-based vPR methods.
This section provides a detailed overview of the dataset, covering its collection methodology, geographic scope, temporal coverage, and suitability for training aerial vPR models.


\begin{figure*}[]
    \centering
    \includegraphics[width=\linewidth]{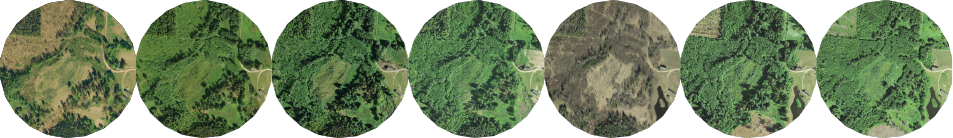} 
    \caption{Examples of images taken from the same location across different years and seasons. The temporal depth and diversity of the dataset are illustrated, highlighting changes due to seasonal variations and environmental transformations.}
    \label{fig:temporal_examples}
\end{figure*}

\begin{figure}[]
    \centering
    \begin{subfigure}[t]{0.45\columnwidth}
        \centering
        \includegraphics[width=\linewidth]{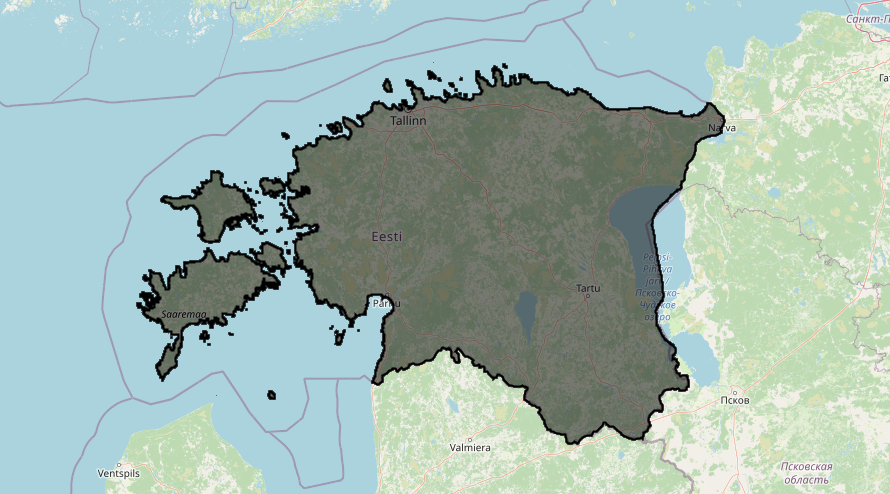} 
        \caption{Boundaries polygon of Estonia}
        \label{fig:subfig1}
    \end{subfigure}
    \hfill
    \begin{subfigure}[t]{0.45\columnwidth}
        \centering
        \includegraphics[width=\linewidth]{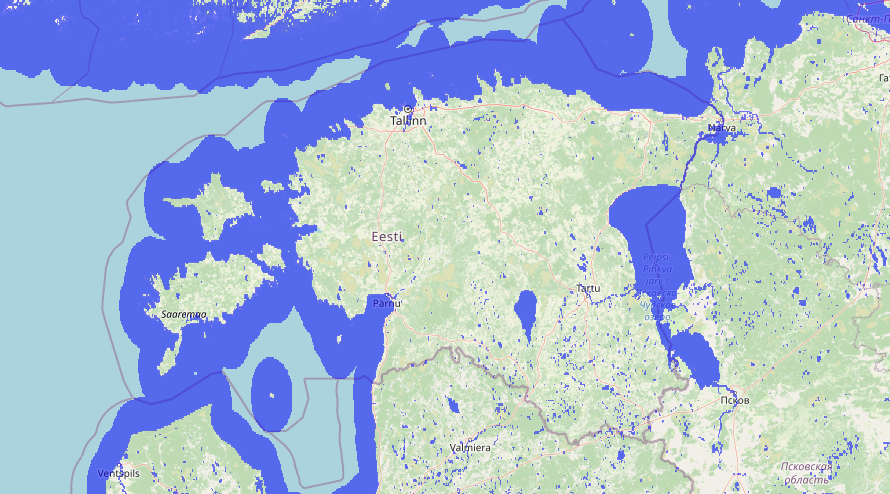} 
        \caption{Global water map.}
        \label{fig:subfig2}
    \end{subfigure}
    \\
    \begin{subfigure}[b]{0.4\columnwidth}
        \centering
        \includegraphics[width=\linewidth]{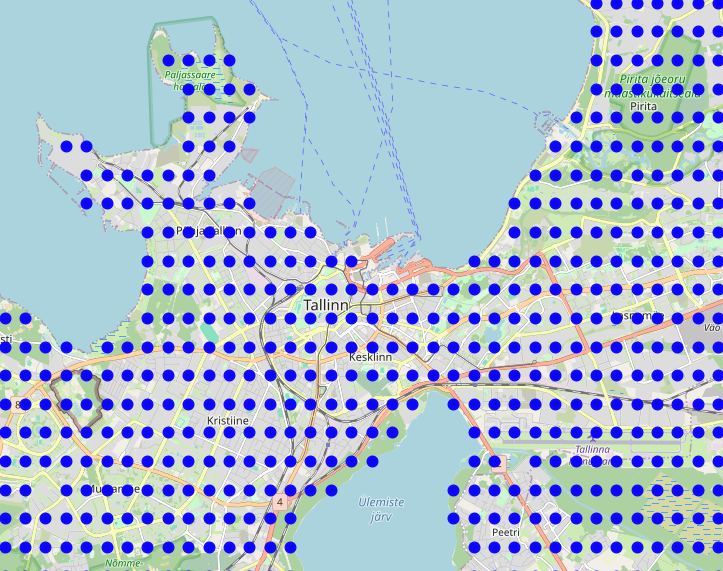} 
        \caption{Example of the sampled locations.}
        \label{fig:subfig3}
    \end{subfigure}
    \hfill
    \caption{Examples of images from the LASED dataset illustrating temporal diversity and geographic variety across Estonia.}
    \label{fig:multi_image_figure}
\end{figure}

\subsection{Data Collection}

The imagery in LASED was captured by the Estonian Land Board, providing nationwide coverage through yearly sampled orthophotos. The original data were accessed in a systematic manner using the Google Earth Engine platform \cite{gorelick_google_2017}, creating a uniform and curated dataset. The captured orthophotos feature a pixel size ranging from 10-16cm for densely populated areas, to 20-40cm for the rest of the country.

To ensure a systematic and uniform dataset, we utilized the country’s boundary polygon\footnote{Global Administrative Unit Layers, Food and Agriculture Organization of the United Nations} to define the sampling area. Using a constant interval in latitude and longitude, we systematically sampled locations across the entire mainland of Estonia. To avoid including images with no useful information, such as those capturing large bodies of water, the sampled locations were filtered using a high-resolution map of global surface water~\cite{pekel_high-resolution_2016}, as shown in Fig. \ref{fig:multi_image_figure}.

\subsection{Dataset Structure} \label{dataset_structure}

Each sampled location represents a unique place, with no overlap between neighboring locations. For a given place, multiple images were captured from a different time frame, ranging from 2011 to 2021, depicting the same scene with pixel-to-pixel correspondence, reflecting changes over time due to seasonal variations and environmental transformations. This approach ensures clear place separation while capturing the temporal evolution of each location.

To address the inherent rotational ambiguity in aerial vPR, the dataset uses circular image sampling instead of standard rectangular cropping. This allows images to be freely rotated without introducing interpolation noise or empty regions. During training, different rotated versions of the same location are generated by rotating the circular image before center-cropping into a rectangle. This process naturally introduces variation into the training data, helping the model to learn more robust representations while reducing the risk of over-fitting during the training phase.
Fig. \ref{fig:temporal_examples} presents examples of images taken from the same location across different years and seasons, showcasing environmental changes over time.

\subsection{Image Properties}
Each image in LASED is circular, with a uniform diameter of 500 pixels, corresponding to a ground area of approximately 400 meters in diameter. To ensure consistency throughout the dataset, all images are aligned to the same orientation. Additionally, each image is accompanied by metadata specifying its precise geographic location and the year it was captured. 

\subsection{Testing Dataset}
Despite the scale and diversity of the proposed dataset, its focus on a single country may limit its ability to evaluate generalization across different geographic and environmental domains. To address this concern, we introduce LASED-Test, a complementary testing dataset sourced from the canton of Valais in Switzerland\footnote{Sourced from the Federal Office of Topography swisstopo}.

LASED-Test was generated using the same systematic methodology as LASED, providing consistency in structure and sampling. It comprises approximately 60 thousand images representing approximately 30 thousand unique locations. The orthophotos were captured in 2017 and 2020, offering a three-year temporal gap that introduces significant changes in environmental conditions. This dataset emphasizes Switzerland’s mountainous landscapes, contrasting with the relatively flat terrain of Estonia, and includes diverse features such as alpine regions and valleys. The orthophotos feature a ground resolution of 10 cm in plains and valleys and 25 cm over the Alps, ensuring detailed representation across varied terrains.

To further evaluate model robustness, we apply fixed geometric transformations to LASED-Test. These transformations simulate real-world UAV operational conditions, such as varying orientations and altitudes, making the testing environment more challenging. Together, the geographic and temporal diversity of LASED-Test provides a robust basis for evaluating the flexibility and generalization capabilities of aerial vPR models.

\section{Addressing UAV Challenges - Rotation}

Similarly to any specific application, UAV vPR poses several unique challenges. One prominent issue, in contrast to most ground-based applications where images are typically upright, is the rotational ambiguity inherent to aerial imagery. To address this, we introduce for the first time in the vPR domain the use of steerable CNNs \cite{weiler_learning_2018}.

\subsection{Equivariance}
Equivariance refers to the property of a model, where the application of a specific transformation to the input results in a corresponding transformation in the output. 
More formally, a function \( f : X \to Y \) is equivariant with respect to a group \( G \) if:

\begin{equation}
f(g\cdot x) = g \cdot f(x), \quad \forall g \in G, \, x \in X,
\end{equation}
where $g \cdot$ denotes a group action in the corresponding space.

Traditional CNNs are inherently translation equivariant, meaning they are well suited for detecting features regardless of their position within an image. However, they lack rotational equivariance, which they need to learn from the data, reducing generalization. 

\begin{figure}[]
    \centering
    \includegraphics[width=\linewidth]{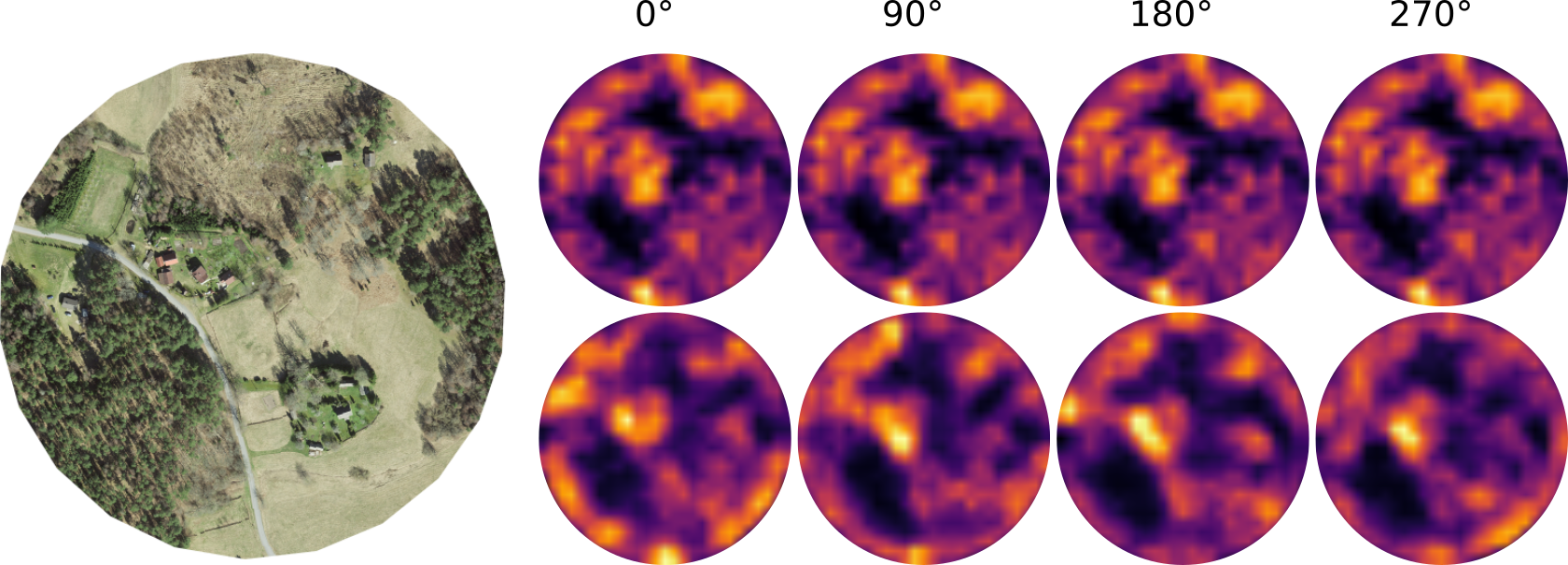} 
    \caption{Visual comparison of feature maps generated by a steerable Convolutional Neural Network (CNN) (top row) and a standard CNN (bottom row) when encoding the same image at different orientations. All feature maps are rotated back to a common reference frame for visualization, highlighting the steerable CNN's ability to maintain consistent representations across orientations. }
    \label{fig:feature_maps}
\end{figure}

Steerable CNNs overcome this challenge by incorporating rotational equivariance directly into their architecture. This allows the network to efficiently recognize features at multiple orientations without directly learning it, by utilizing steerable filters. By leveraging these networks, we enhance the robustness of feature representations and improve the overall vPR performance. As illustrated in Fig. \ref{fig:feature_maps}, the feature maps produced by a steerable Convolutional Neural Network (CNN) remain stable across different input orientations, whereas those from a standard CNN exhibit noticeable variations, highlighting the advantage of equivariant representations for aerial vPR.

\subsection{Steerable Filter}
The design of a steerable filter, is based on the idea that a filter of arbitrary orientations can be generated by linearly combining a set of some basis functions $\{\psi_{q=1}^{Q}\}$. More specifically:


\begin{equation}
    \Psi^{\theta}(x) = \sum^{Q}_{q=1}{k_{q}(\theta)\psi_{q}(x)},
\end{equation}
where $\Psi : \mathbb{R}^2 \to \mathbb{R}$ is the steerable filter 
$\forall \theta \in(-\pi,\pi]$, and 
$k_q$ are interpolation functions~\cite{freeman_design_1991,weiler_learning_2018}. 

\subsection{Implementation Details} \label{implementation}
The steerable convolutional neural network backbone is constructed using steerable modules implemented in \cite{weiler_learning_2018,cesa_program_2021}. These modules are incorporated into the ResNet architecture, making the network equivariant to rotations.

Unlike standard convolutional networks, steerable CNNs produce feature maps with an additional orientation dimension, resulting in a tensor of shape $(C, N, H, W)$, where $C$ is the number of feature channels, $N$ represents the number of discrete orientations (e.g., 8 for $C_8$ and 4 for $C_4$), and $(H, W)$ denote the spatial dimensions. To achieve rotation invariance, we first apply orientation pooling across the $N$ dimension, aggregating information from all orientations and reducing the feature map to $(C, H, W)$. This ensures that the final representation is invariant to rotations while retaining discriminative features. Finally, spatial pooling is applied over $(H, W)$, producing a compact vector representation suitable for vPR tasks.

In our experiments, we tested steerable ResNet variants with cyclic groups $C_8$ and $C_4$, corresponding to equivariance under discrete rotations of $45^{\circ}$ and $90^{\circ}$, respectively. These models, denoted as \texttt{sresnet50c8} and \texttt{sresnet50c4}, incorporate steerable convolutional layers that maintain equivariance to their respective rotation groups. The choice of $C_8$ allows for finer rotational sensitivity, while $C_4$ provides a coarser but more computationally efficient alternative. This setup ensures a balance between capturing rotational variations and maintaining efficiency for aerial vPR tasks.

\section{Experiments}
The experiments carried out in this work aim to evaluate the effectiveness of both the proposed steerable ResNet architecture and the LASED in addressing the challenges of aerial vPR. The evaluation focuses on two key aspects: assessing the performance of the steerable network in comparison to state-of-the-art vPR methods, and analyzing the impact of LASED as a training dataset compared to existing aerial vPR datasets.

To achieve this, we conduct a series of experiments across multiple datasets and configurations, comparing the performance of the steerable network against other approaches and examining the generalization capabilities of models trained on LASED versus other datasets. These evaluations are performed under various conditions, including diverse geographic regions, temporal variations, and environmental changes, to provide a comprehensive analysis of the proposed approach.

The following subsections provide a detailed description of the experimental setup, evaluation metrics, and the obtained results.

\subsection{Experimental Setup}
\subsubsection{Datasets}
To comprehensively evaluate the proposed approach, we conducted experiments across multiple aerial vPR datasets, including our LASED dataset and several publicly available datasets widely used in the field. The datasets utilized in this study are as follows:

\begin{itemize} \item \textbf{LASED (Ours):} A large-scale aerial dataset covering the entirety of Estonia, comprising of $~1M$ images captured over a decade, featuring over $170k$ unique places. It offers multi-season and long-term temporal coverage, making it a valuable resource for training robust aerial vPR models. 
\item \textbf{LASED-Test (Ours):} A complementary dataset focusing on the canton of Valais in Switzerland, consisting of $~60k$ images captured across a three-year span. It introduces geographic diversity with challenging mountainous terrain, providing an ideal benchmark for evaluating generalization across different regions. 

\item \textbf{VPAir~\cite{schleiss_vpair_2022}:} A publicly available aerial vPR dataset containing high-altitude images spanning urban, grassland, and forest environments along its more than one hundred kilometers long trajectory over a region near the city of Bonn, Germany. The provided queries were recorded using a downward facing camera on board a light aircraft in 2020, while the reference and the distractor images are publicly available orthophotos captured between 2019 and 2021. A match is considered positive if it is within three image range. 

\item \textbf{ALTO~\cite{cisneros_alto_2022}:} A medium-scale dataset, part of the general place recognition competition (round 2) for International Conference on Robotics and Automation (ICRA) 2022. It features two trajectories of a helicopter flying over Ohio and Pennsylvania, along with publicly available orthoimagery, covering diverse environments, including forests, urban areas, rural regions, rivers, lakes, and grasslands. The dataset provides training, validation, and testing sets. However, due to the unavailability of the official evaluation server and the absence of ground-truth data for the testing images, we used the validation partition for evaluation and augmented it with the testing images as distractors.

\item \textbf{SUES-200~\cite{zhu_sues-200_2023}:} A large dataset comprising of 200 locations around the Shanghai University of Engineering and Science. Its structure consists of one satellite image per location and multiple drone images circling that location at different heights. For our evaluation, we utilize the $300m$ altitude flights test set.

\item \textbf{University-1652~\cite{zheng_university-1652_2020}:} It features $1652$ discrete locations around 72 universities and it consists of three modalities, $i.e.,$ ground-based images, synthetic UAV views, as well as satellite-views. It utilizes the Google Map service to source the satellite-views and Google's Earth 3D models to simulate the UAV flights, spiraling around each location while descending from $256m$, to $121.5m$. Unlike strictly nadir (top-down) aerial datasets, the UAV views in University-1652 are captured with a forward-tilted camera angle, resulting in oblique perspectives that include both the ground and horizon. This introduces additional viewpoint variation, making the dataset more challenging. For this evaluation, we use the $query\_drone$ and $gallery\_satellite$ data from the test partition.

\item \textbf{AerialVL~\cite{he_aerialvl_2024}:} It consists of images captured from a flying drone combined with satellite views at three different heights, used for reference. The dataset spans an area around $20 km^2$, with both urban and rural zones as well as multiple lakes and rivers. For testing purposes, we leverage the highest satellite-view altitude (level 3) combined with the drone-captured images, reaching $22,916$ images in total. Furthermore it provides an extra set of images for training collected from satellite-view imagery covering an area of $54km^2$ with $22,000$ images. A positive match is considered when the two images are within a $50m$ radius.

\end{itemize}
As we can observe AerialVL and Alto are the most relatable datasets to ours, covering several different area types and providing data for both evaluation and testing purposes. To that end, we select them as benchmarks for the comparative evaluation of training datasets. Unlike the proposed dataset, their structure is in the form of database-query. During training, we treat each query as a separate place, pairing it with the corresponding database images as positives. Negative pairs are mined from the remaining images. Although this approach introduces a small amount of noise due to overlap between consecutive queries, it still proves more effective than using only positive pairs.  

\subsubsection{Evaluated Methods and Metrics}
To assess the effectiveness of incorporating steerable CNNs in aerial vPR, we compare their performance against top-performing state-of-the-art methods, utilizing the widely adopted ResNet as a backbone. The evaluated methods include established aggregation techniques such as ConvAP~\cite{ali-bey_gsv-cities_2022}, CosPlace~\cite{berton_rethinking_2022}, MixVPR \cite{ali-bey_mixvpr_2023}, and Generalized Mean Pooling (GeM)~\cite{radenovic_fine-tuning_2019}, which are commonly used in vPR tasks. In addition to the state-of-the-art, we also benchmark a baseline method, using the standard ResNet architecture as a backbone and the same spatial aggregation technique as the one utilized by the proposed steerable network, described in Subsection \ref{implementation}. 

For a comprehensive evaluation, we utilize the commonly adopted metric of \textit{Recall@N}, where the percentage of correctly identified locations is measured at top 1, 5, and 10 results. This metric provides insights into the effectiveness of different models in ranking the correct place among the top retrieved candidates, reflecting their practical utility in real-world scenarios.

\subsubsection{Training Details}
All models were trained following a standardized pipeline to ensure a fair comparison across different methods and datasets. Training was conducted for 30 epochs, with all images resized to a fixed resolution of 256 × 256 pixels to maintain consistency and improve computational efficiency, utilizing the state-of-the-art multi-similarity loss with its online mining scheme~\cite{wang_multi-similarity_2019}. The training was conducted on 4 A10G GPUs and 48 vCPUs with 192 GB of memory. Additionally, to further increase the training speed, we make use of half precision.

To enhance generalization, data augmentation techniques were applied uniformly across all training datasets. Random rotations were used to account for UAV orientation variability, while the automated augmentation method RandAugment~\cite{cubuk_randaugment_2019} introduced additional appearance variations, improving robustness to environmental changes. For LASED, the rotation and resizing scheme followed the process described in Subsection \ref{dataset_structure}.

\begin{table*}[!ht]
    \centering
    \scriptsize
    \centering
    \begin{tabular}{lcllllll}
        ~ & ~ & AerialVL & Alto & SUES200 & University1652 & VPair & LASED\_test \\ \hline
        method & training dataset & R@1/R@5/R@10 & R@1/R@5/R@10 & R@1/R@5/R@10 & R@1/R@5/R@10 & R@1/R@5/R@10 & R@1/R@5/R@10 \\ \hline

cosplace    & lased            & 49.81/74.31/80.35                & 12.59/57.19/79.22            & 99.00/100.00/100.00             & 29.86/49.90/58.87                      & 79.93/88.54/91.09             & 57.92/65.49/68.20                   \\
convap      & lased            & 57.07/75.97/80.72                & 11.28/54.93/84.62            & 98.50/99.00/99.00               & 12.99/27.10/35.61                      & 42.76/60.53/69.55             & 28.25/40.43/45.92                   \\
gem         & lased            & 37.38/61.53/69.40                & 10.51/47.86/72.57            & 97.00/98.50/99.00               & 19.03/36.09/45.42                      & 68.40/82.08/86.99             & 44.86/54.59/58.27                   \\
mixvpr      & lased            & 50.40/69.69/75.74                & 18.17/72.51/88.78            & 98.00/99.00/99.50               & 26.52/45.79/54.98                      & 69.51/80.49/84.26             & 37.74/49.29/53.95                   \\
baseline    & lased            & 44.47/71.09/77.94                & 9.80/48.99/73.16             & 98.50/99.50/99.50               & 25.20/44.64/54.28                      & 75.68/86.51/89.62             & 53.57/62.56/65.68                   \\
sresnet50c4 & lased            & 62.76/81.49/85.67                & 19.66/75.36/93.17            & 99.50/100.00/100.00             & 38.30/58.01/66.43                      & 81.56/87.84/89.84             & 60.01/67.57/70.30                   \\
sresnet50c8 & lased            & 64.61/81.54/85.35                & 20.55/81.47/95.90            & 98.50/99.00/99.00               & 38.68/58.22/66.54                      & 81.86/88.32/90.76             & 64.50/71.36/73.76                   \\
sresnet18c4 & lased            & 60.91/79.55/84.31                & 15.80/68.82/90.08            & 99.50/100.00/100.00             & 34.70/55.31/64.33                      & 75.87/85.37/88.43             & 57.85/66.19/69.29                   \\
sresnet18c8 & lased            & 61.13/79.36/84.02                & 17.04/72.33/90.50            & 98.00/98.50/98.50               & 27.85/47.35/56.64                      & 75.28/84.37/87.92             & 59.70/67.95/70.73                   \\
 \hline            
cosplace    & aerialvl         & 27.09/49.30/55.88                & 3.56/14.37/23.81             & 94.00/98.50/99.00               & 20.31/38.93/48.78                      & 43.13/61.79/68.26             & 24.35/34.88/39.71                   \\
convap      & aerialvl         & 51.03/68.16/72.45                & 9.68/38.12/58.55             & 96.50/99.00/99.00               & 15.98/28.43/35.12                      & 21.36/31.37/37.44             & 12.05/18.13/21.45                   \\
gem         & aerialvl         & 25.08/45.89/54.27                & 5.05/23.34/36.34             & 83.00/88.50/93.50               & 17.55/32.95/41.35                      & 24.02/40.84/48.85             & 14.43/23.78/28.28                   \\
mixvpr      & aerialvl         & 35.34/53.63/60.64                & 5.23/24.52/37.35             & 94.00/97.50/99.00               & 13.94/28.92/37.88                      & 26.64/44.97/53.55             & 11.43/20.16/24.96                   \\
baseline    & aerialvl         & 23.00/43.71/52.17                & 4.39/20.13/32.19             & 96.00/98.00/98.50               & 18.08/35.64/45.21                      & 41.54/58.57/66.56             & 23.03/33.46/38.21                   \\
sresnet50c4 & aerialvl         & 37.94/62.84/69.22                & 4.45/25.95/39.55             & 97.00/99.00/99.00               & 23.77/42.98/52.55                      & 52.33/67.52/74.50             & 33.48/44.12/48.42                   \\
sresnet50c8 & aerialvl         & 42.54/68.98/74.93                & 6.83/29.39/48.34             & 98.00/99.00/100.00              & 21.93/40.01/49.21                      & 56.39/72.28/78.16             & 39.21/49.63/53.69                   \\
sresnet18c4 & aerialvl         & 43.29/68.11/74.43                & 3.38/17.28/25.71             & 96.50/98.50/98.50               & 24.05/42.70/52.05                      & 47.89/64.15/70.69             & 33.97/44.60/49.34                   \\
sresnet18c8 & aerialvl         & 34.85/58.53/64.44                & 4.28/17.93/26.60             & 96.50/99.50/100.00              & 19.92/37.63/47.01                      & 48.45/66.19/72.65             & 37.64/48.30/52.51                   \\
 \hline         
cosplace    & alto             & 25.35/47.10/55.33                & 6.65/31.71/50.18             & 91.50/97.50/98.00               & 12.57/26.31/34.30                      & 35.62/55.65/63.86             & 15.15/24.91/29.55                   \\
convap      & alto             & 41.57/63.64/71.44                & 11.76/47.03/70.84            & 94.50/97.00/98.00               & 9.95/19.88/26.07                       & 17.04/28.57/35.55             & 8.36/14.04/17.24                    \\
gem         & alto             & 14.85/31.62/40.76                & 6.53/37.59/59.74             & 70.50/79.50/86.00               & 10.42/23.56/31.79                      & 19.99/37.40/46.45             & 12.37/21.25/25.93                   \\
mixvpr      & alto             & 22.13/37.49/44.92                & 11.88/41.09/55.23            & 82.00/91.50/95.50               & 7.13/16.45/23.00                       & 19.62/35.48/45.01             & 8.03/16.12/20.89                    \\
baseline    & alto             & 21.24/41.69/50.20                & 6.12/25.59/43.71             & 91.00/96.50/98.00               & 12.80/26.52/34.77                      & 38.36/57.61/66.30             & 16.76/26.37/31.07                   \\
sresnet50c4 & alto             & 37.10/56.63/62.61                & 10.45/42.76/64.67            & 97.50/99.50/99.50               & 17.60/32.28/40.49                      & 46.30/62.56/69.81             & 26.91/37.31/41.98                   \\
sresnet50c8 & alto             & 45.69/66.73/72.62                & 11.64/51.13/73.28            & 97.50/98.50/99.50               & 25.51/44.67/53.98                      & 47.78/63.78/70.25             & 29.68/39.98/44.14                   \\
sresnet18c4 & alto             & 34.82/57.74/65.27                & 9.32/38.84/58.43             & 96.50/98.50/99.00               & 18.40/35.93/45.20                      & 36.07/54.80/61.09             & 23.15/33.83/38.68                   \\
sresnet18c8 & alto             & 37.71/59.89/66.53                & 7.13/32.72/50.95             & 94.00/98.00/99.50               & 16.17/32.57/41.68                      & 33.89/53.40/60.53             & 27.91/38.97/43.77                  
    \end{tabular}

    \centering
\caption{Performance comparison of different models trained on LASED, AerialVL, and Alto datasets. Performance is evaluated using recall at top 1, 5, and 10 retrievals across multiple test datasets. The dimensions of the produced representation vectors of all methods are 512 except GeM where they are 2048.}
\label{tab:results}
\end{table*}

\subsection{Results}
\subsubsection{Evaluation of the LASED Dataset}
This experiment aims to evaluate the impact of the LASED dataset, when used for training, on model performance and its ability to generalize across diverse environments. To achieve this, multiple models were trained separately using LASED, AerialVL, and Alto datasets, and their performance was assessed on the same set of test datasets. The results, presented in Table \ref{tab:results}, emphasize the importance of large and diverse training datasets in achieving superior generalization performance. Models trained on LASED consistently outperform those trained on AerialVL and Alto across all test datasets, highlighting the advantages of LASED’s extensive geographic coverage, larger scale, and long-term temporal span.

\begin{figure}[]
    \centering
    \includegraphics[width=\linewidth]{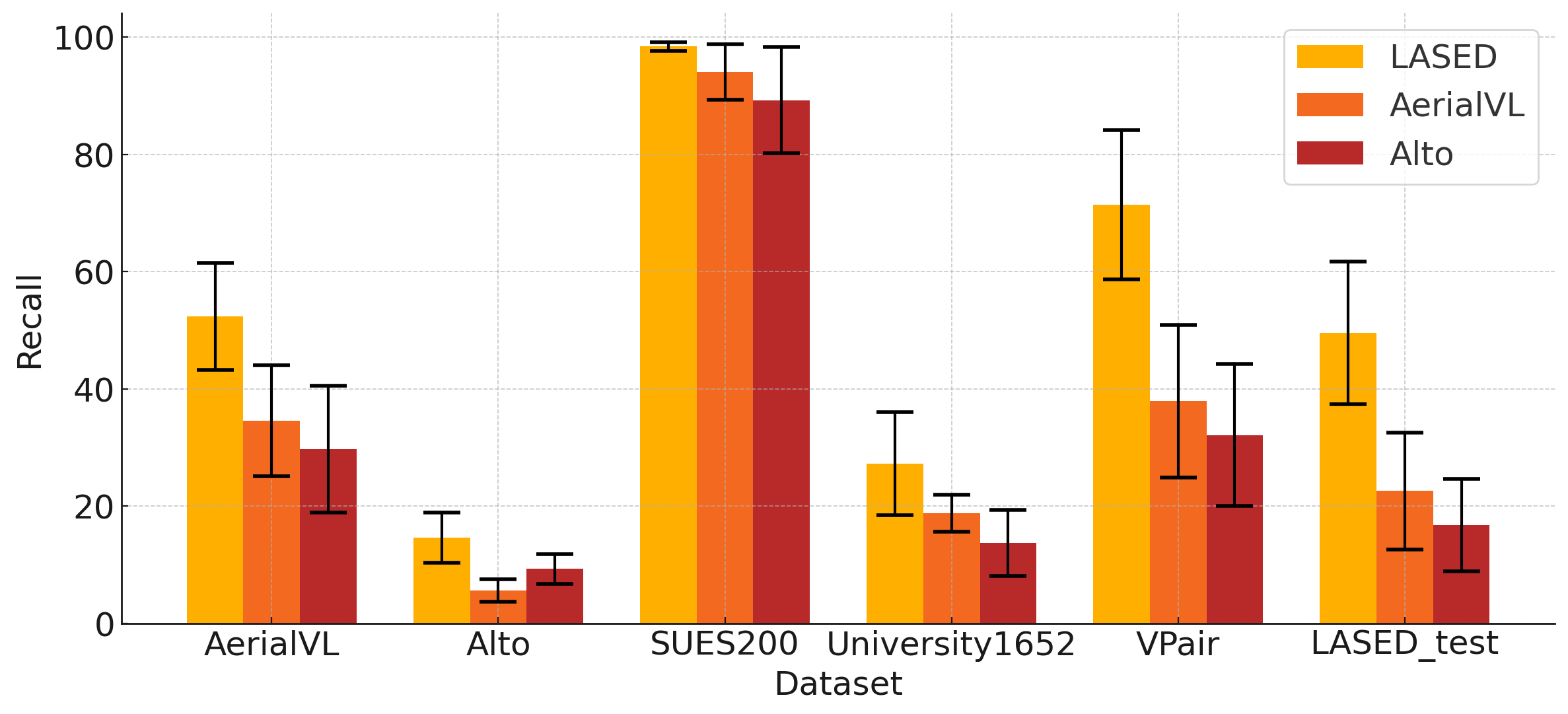} 
    \caption{Comparison of the average $Recall@1$ and standard deviation for all models in Table \ref{tab:results} across all test datasets, when trained on LASED, AerialVL, and Alto. This highlights the impact of dataset scale and diversity on model generalization and recognition performance.}
    \label{fig:dataset_comp}
\end{figure}

Fig. \ref{fig:dataset_comp} summarizes the comparison by presenting the average $Recall@1$ as well as the standard deviation across all test datasets for the models in Table \ref{tab:results}, trained on all three training datasets, offering insight into the influence of training data on recognition performance, despite the utilized model architecture.

\subsubsection{Evaluation of the Steerable Network}
Through this benchmark, we assess the effectiveness of the proposed steerable network in enhancing aerial vPR performance by addressing the challenge of rotational variance inherent in UAV imagery. The steerable network is compared against state-of-the-art architectures using identical training conditions and datasets to ensure a fair evaluation.

The evaluation is performed on multiple test datasets that capture diverse geographic and temporal conditions. The results, summarized in Table \ref{tab:results}, demonstrate that the steerable network consistently outperforms traditional convolutional models, achieving higher recall in various scenarios. More specifically, $sresnet50c8$ on average performs 12\% better than the best scoring non-steerable architecture when trained on LASED, 9\% when trained on AerialVL, and 35\% when trained on Alto dataset. 

An interesting observation is the performance of ConvAP, which, despite under-performing on most datasets, achieves above-average results on AerialVL and Alto. This can be attributed to its spatial aggregation technique, which favors datasets with minimal rotational variance. In contrast, its poor performance on datasets with higher rotational variance highlights a limitation not observed in \cite{ali-bey_gsv-cities_2022}, where ConvAP outperforms CosPlace in ground-based vPR setups. This suggests that while ConvAP’s focus on spatial consistency is effective in environments with limited rotational variance, it hinders performance in aerial datasets where rotational invariance is crucial.

These results highlight a fundamental difference between aerial vPR and traditional ground-based vPR. While methods like ConvAP demonstrate strong performance in ground-based scenarios, their reliance on spatial consistency limits their effectiveness in aerial environments where rotational variance is a dominant factor. The consistent performance improvements observed with the steerable network further emphasize the need for architectures that explicitly address rotational ambiguity, a challenge less pronounced in ground-based applications. 

\begin{table}[t]
\centering
\caption{Comparison of steerable and standard convolutional models trained with and without rotation augmentations, evaluated on the LASED\_test dataset.}
\label{tab:no_rotation}
\begin{tabular}{l|c|c}
\hline
method      &           w/o rotation augmentations & w rotation augmentations \\
            &         R@1/R@5/R@10 & R@1/R@5/R@10\\ \hline
cosplace               & 14.69/18.39/20.30  & 57.92/65.49/68.20  \\  
sresnet50c4            & 15.57/21.34/24.26  & 60.01/67.57/70.30 \\  
sresnet50c8            & 22.41/30.86/34.82  & 64.50/71.36/73.76\\   
       \hline
\end{tabular}
\end{table}

Table \ref{tab:no_rotation} presents a direct comparison between the performance of steerable CNNs and standard CNNs with CosPlace, both trained on LASED without explicit rotation augmentations (aside from limited rotation introduced by RandAugment). The results reveal that while steerable CNNs still outperform standard CNNs, their performance significantly drops without sufficient rotation augmentations. This suggests that although steerable networks inherently encode rotational equivariance, in practice, they still benefit from being exposed to rotated examples during training. This highlights the importance of combining rotation-equivariant architectures with appropriate data augmentation to fully exploit their potential in aerial vPR tasks.


\subsection{Number of Model Parameters}
\begin{figure}[]
    \centering
    \includegraphics[width=\linewidth]{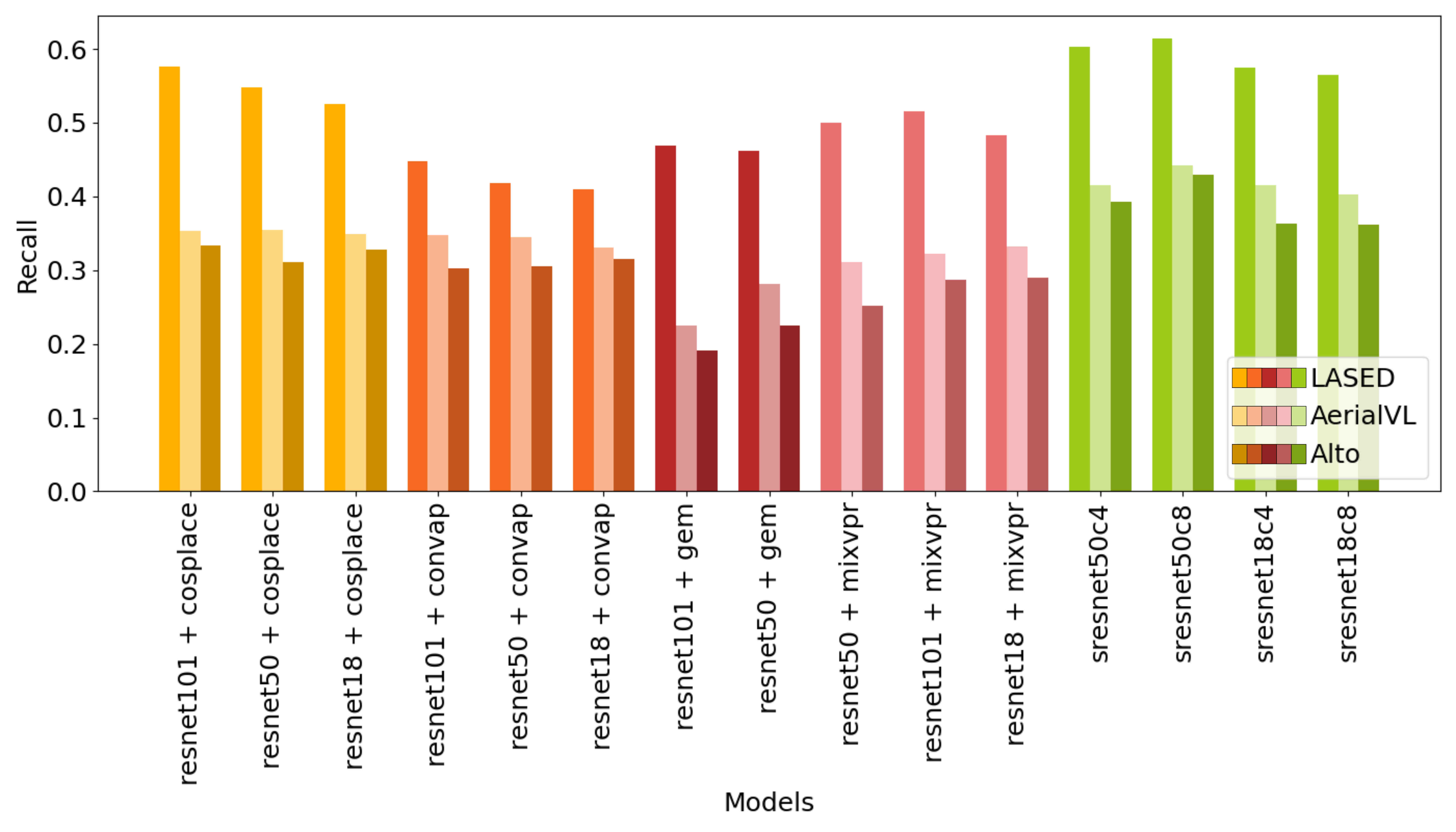} 
    \caption{Average $Recall@1$ across all test datasets for different backbone sizes. Brighter-colored bars correspond to models trained on LASED, faded-colored bars represent training on AerialVL, while bolder-colored bars indicate models trained on Alto. Models trained on LASED benefit more from increased capacity, whereas those trained on AerialVL and Alto show minimal or even negative gains, emphasizing the importance of training data diversity and scale.}
    \label{fig:model_size}
\end{figure}

Model capacity, often reflected in the number of parameters within the backbone architecture, plays a crucial role in determining performance. Understanding how model size influences recall is important for optimizing vPR systems across different scenarios. Fig. \ref{fig:model_size} illustrates the average performance of each method across all test datasets when using different backbone sizes. As expected, larger models generally perform better than their smaller counterparts. However, this trend varies significantly depending on the training dataset. When trained on AerialVL or Alto, the performance gain from increasing model size is minimal, and in some cases, even negative. In contrast, models trained on LASED exhibit substantial performance improvements as the backbone size increases.

These results indicate that the effectiveness of larger models is closely linked to the diversity and scale of the training data. In scenarios where the training data are limited or lack diversity, larger networks may not fully realize their potential and could even underperform. Conversely, when trained on large, diverse datasets like LASED, larger models can leverage the richness of the data to significantly outperform smaller architectures. This highlights the importance of aligning model capacity with the characteristics of the training data to achieve optimal performance.

\subsection{Number of Dimensions}
\begin{figure}[]
    \centering
    \includegraphics[width=\linewidth]{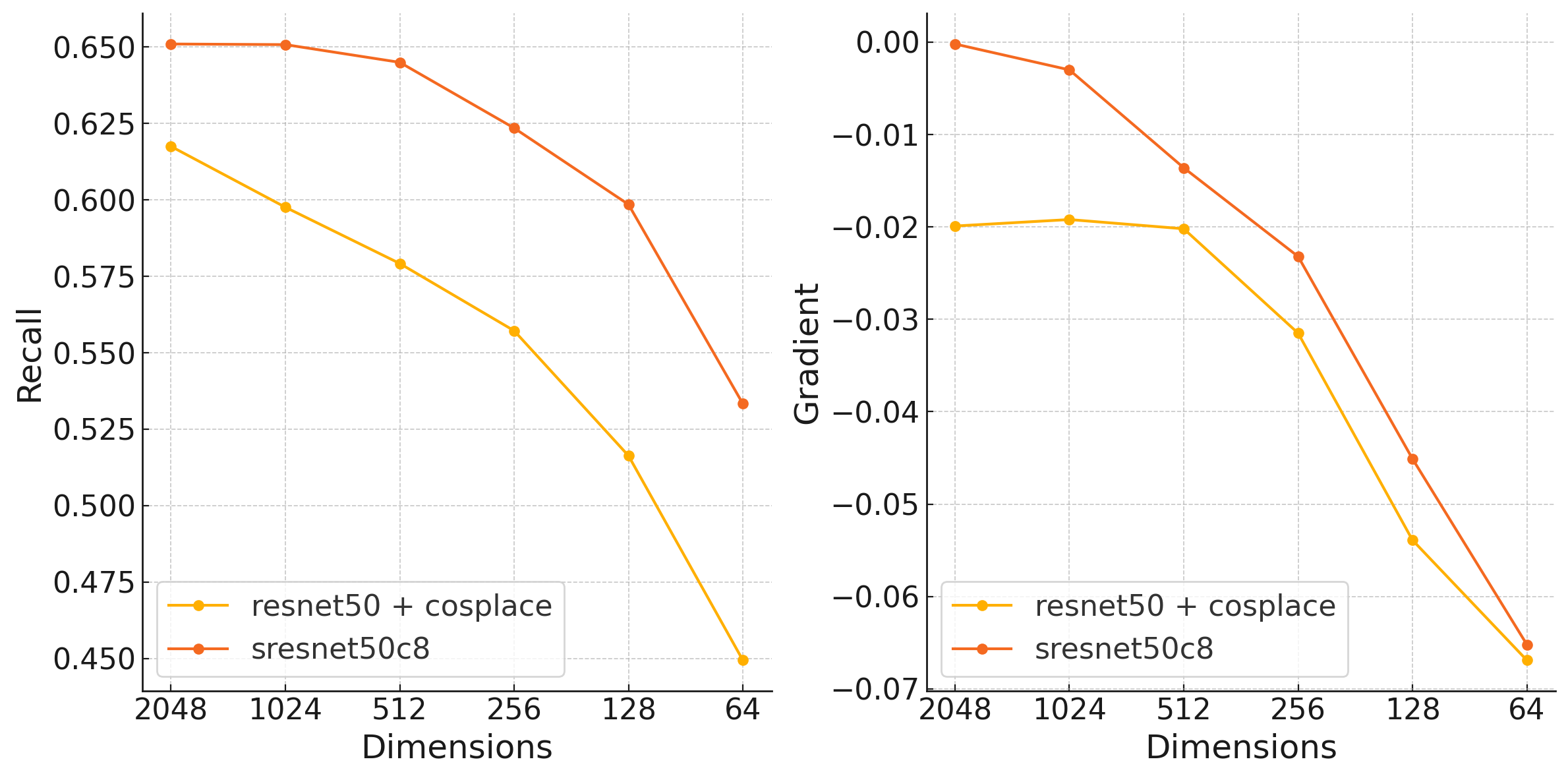} 
    \caption{Impact of feature dimensionality on vPR performance for CosPlace and sresnet50c8, evaluated on LASED-Test. The left plot shows the $Recall@1$ as the feature vector dimensionality decreases, while the right plot presents the gradient of the recall curves, illustrating the rate of performance decline for each method.}
    \label{fig:dimensions}
\end{figure}

The dimensionality of the feature vector plays a crucial role in visual place recognition, influencing both recall and computational efficiency. To analyze this, we retrained the networks with different output dimensions and evaluated their performance on the LASED-Test dataset. Fig. \ref{fig:dimensions} presents the performance trends for two representative methods, CosPlace and $sresnet50c8$, showing how recall changes as feature dimensionality is reduced. The gradient of each line highlights the rate of performance decline, offering insight into how each model retains discriminative power at lower dimensions.

These findings underscore the importance of choosing the right feature dimensionality, as performance depends on the model’s capacity to maintain robust representations. The steerable design of $sresnet50c8$ supports this by effectively preserving discriminative features even when the dimensionality is reduced.

\subsection{Runtime Analysis}
To further assess the practical feasibility of the proposed approach, we measured the average encoding time per image for both the steerable network and the baseline model on the LASED-Test dataset. All experiments were conducted on a system equipped with an Intel i7 14700K processor, 64 GB RAM, and an RTX 4070 Ti GPU, using batch size equal to 1. The results presented in Table \ref{tab:runtime} indicate that while steerable networks introduce additional computational overhead compared to the baseline ResNet model, the increase remains within an acceptable range for real-time applications. Specifically, $sresnet50c8$ exhibits the highest encoding time at $14.1ms$ per image, whereas the baseline model achieves significantly lower latency at $2.3ms$ per image. This trade-off highlights the balance between improved robustness to rotational variance and computational efficiency. In particular, smaller steerable architectures, such as $sresnet18c4$ reduce the time gap, while still outperforming most ResNet-based non-steerable methods (Table \ref{tab:results}).

\begin{table}[!ht]
\centering
\caption{Average encoding time per image on the LASED-Test dataset.}
\label{tab:runtime}
\begin{tabular}{l|c|c}
\hline
method        & time ($ms$) & std ($ms$) \\ \hline
sresnet50c8            & 14.141           & 9.740  \\       
sresnet50c4            & 7.833            & 1.946    \\     
sresnet18c8            & 8.688            & 7.492   \\      
sresnet18c4            & 5.528            & 6.546  \\       
baseline               & 2.280            & 1.221   \\         \hline
\end{tabular}
\end{table}

\subsection{Discussion}
This work demonstrates how large-scale, structured datasets and rotation-equivariant architectures improve aerial vPR, making it more scalable and robust for real-world UAV applications.

\subsubsection{Impact of LASED on Aerial vPR}
Our results highlight the critical role of LASED in improving model generalization. Models trained on LASED consistently outperform those trained on AerialVL and Alto across all test datasets. This improvement is expected, given LASED’s extensive geographic coverage, diverse environmental landscapes (urban, rural, forests, coastlines etc.), and decade-long temporal span. Unlike other aerial datasets that are either geographically limited or lack long-term temporal variation, LASED provides a more representative training ground, enabling models to handle changing seasons, evolving infrastructure, and natural transformations over time.
\subsubsection{Advantage of Steerable Networks}
Our results demonstrate that steerable networks effectively address the rotation variance challenge inherent in aerial vPR. Unlike standard CNNs, which need to learn rotational patterns from data, steerable networks inherently encode equivariant representations, allowing them to generalize better across varying UAV orientations. This results in more stable feature representations and improved recognition performance across diverse test datasets.

Additionally, steerable networks achieve this performance improvement without requiring larger feature dimensions. As shown in our experiments, $sresnet50c8$ consistently outperforms non-steerable architectures, even at reduced feature dimensions, rendering it a more memory-efficient choice for large-scale deployment. This characteristic is particularly important for UAV applications, where computational efficiency and storage constraints are critical factors.

\subsubsection{Scalability of Aerial vPR}
A key challenge in deploying aerial vPR for large-scale applications, such as national-scale UAV navigation, is the feasibility of storing and retrieving place descriptors efficiently in memory. Our results show that models trained on LASED achieve higher recall across diverse environments, making them better suited for large-scale mapping. Additionally, reducing feature dimensionality significantly impacts storage requirements, influencing the practicality of nationwide deployment.

To estimate the memory demands of large-scale deployment, we consider an image sampling approach similar to LASED. Assuming each image covers approximately $0.08~km^2$, fully mapping a large country like Germany ($357,000~km^2$) would require around 4.5 million images. With a 512-dimensional half-precision feature vector per image, the total memory needed would be about 4.5 GB.

While this is a simplified estimation, it highlights the feasibility of storing nationwide place descriptors on UAV systems. Steerable networks, which retain higher recall scores at reduced feature dimensions, offer some advantages in minimizing storage needs while maintaining performance. However, even non-steerable architectures trained on LASED show strong results, reinforcing the importance of both high-quality training data and efficient feature encoding in making large-scale aerial vPR practical.

\subsubsection{Computational Tradeoffs}

Balancing high recall with computational efficiency is crucial for real-world UAV applications. While steerable networks improve aerial vPR performance, they come with an increased computational cost compared to standard convolutional architectures. This tradeoff becomes especially important for UAVs with limited onboard processing power.

Our runtime measurements on the LASED-Test dataset confirm this tradeoff. As shown in Table \ref{tab:runtime}, steerable networks take longer to encode an image than the baseline ResNet model. For example, $sresnet50c8$ requires $14.1~ms$ per image, while the baseline model completes the same task in just $2.3~ms$. This additional processing time comes from the extra computations needed to enforce rotation equivariance.

Whether this tradeoff is acceptable depends on the specific application. For UAVs with limited computational resources, the slower encoding speed of steerable networks could be a limiting factor. In such cases, smaller steerable models like $sresnet18c4$ offer a practical compromise, reducing computational load while still outperforming standard architectures in recall.

Ultimately, the choice between accuracy and efficiency depends on deployment constraints. Steerable networks offer improved recall, but their higher computational cost may not always be ideal for real-time applications with strict latency requirements. However, smaller variants like $sresnet18c4$ provide a balanced tradeoff, reducing computational overhead while retaining many of the benefits of rotational equivariance. Selecting the right model depends on the specific needs of the deployment, whether prioritizing accuracy, speed, or a compromise between the two.
\section{Conclusions}
The current work explored the challenges of large-scale aerial vPR and proposed two key contributions: LASED, a large-scale high-altitude vPR dataset with extensive geographic and temporal diversity, and the integration of steerable convolutional networks to improve robustness to rotational variance.

Through extensive benchmarking, we demonstrated that models trained on LASED significantly improve recognition accuracy, outperforming those trained on smaller, geographically constrained datasets. Additionally, our evaluation of steerable networks highlighted their ability to naturally handle aerial image rotations, leading to better generalization across diverse environments. While steerable models introduce some computational overhead, smaller variants offer a practical tradeoff, maintaining strong performance while reducing processing costs. By advancing both data availability and model robustness, this work contributes to scaling aerial vPR for real-world UAV applications.
\bibliographystyle{IEEEtran}
\bibliography{bibliography}

\end{document}